%% file: main_parsing_speech.tex
\documentclass[a4paper]{article}

\usepackage{INTERSPEECH2019}
\usepackage{url}
\usepackage{multirow}
\usepackage{xcolor}
\usepackage{subcaption,enumitem}
\usepackage{booktabs}
\usepackage{comment}

\title{On the Role of Style in Parsing Speech with Neural Models}
\name{Trang Tran$^1$, Jiahong Yuan$^2$, Yang Liu$^2$, Mari Ostendorf$^1$}
\address{
  $^1$Electrical \& Computer Engineering, University of Washington\\
  $^2$LAIX Inc.}
\email{\{ttmt001,ostendor\}@uw.edu, \{jiahong.yuan,yang.liu\}@liulishuo.com}

\begin{document}

\maketitle
\begin{abstract}
The differences in written text and conversational speech are substantial; previous parsers trained on treebanked text have given very poor results on spontaneous speech. For spoken language, the mismatch in style also extends to prosodic cues, though it is less well understood.  This paper re-examines the use of written text in parsing speech in the context of recent advances in neural language processing. We show that neural approaches facilitate using written text to improve  parsing of spontaneous speech, and that prosody further improves over this state-of-the-art result. Further, we find an asymmetric degradation from read vs.\ spontaneous mismatch, with spontaneous speech more generally useful for training parsers. 
\end{abstract}
\input{intro.tex}

\input{models.tex}

\input{data-metrics.tex}

\input{experiments.tex}

\input{conclusion.tex}

\input{thanks.tex}

\bibliographystyle{IEEEtran}

\bibliography{interspeech19}

\end{document}

%% file: intro.tex
\section{Introduction}
Constituency parsing is a well-studied problem in natural language processing, but most state-of-the-art parsers have only been tested on written text, e.g.\ the standard Penn Treebank Wall Street Journal (WSJ) dataset \cite{Marcus1999}. 
These recent neural parsers are commonly formulated as encoder-decoder systems, where the encoder learns the input sentence representation and the decoder learns to predict a parse tree. While input is often represented by word-level features, representation for the output trees varies: 
as a sequence of parse symbols \cite{Vinyals2015},
a set of spans \cite{Stern2017}, 
syntactic distances \cite{Shen2018}, or per-word structure-rich labels \cite{Vilares2018}.
A key characteristic in many of these neural parsers is the recurrent network structure, particularly Long Short-Term Memory networks (LSTMs); however, Kitaev and Klein \cite{Kitaev2018} have shown that a non-recurrent encoder such as the Transformer network introduced in \cite{Vaswani2017} is also capable of encoding timing information through self-attention mechanisms, achieving state-of-the-art parse results on the Treebank WSJ dataset. 
Further, these parsers 
benefit from contextualized information learned from larger external text data, such as ELMo \cite{Peters2018} and BERT \cite{Devlin2018}.

It is not clear that these advances will transfer to speech data, particularly for the different styles of speech. Even when perfect transcripts are available, speech poses many challenges to parsers learned from written text due to the lack of punctuation and case, and the presence of disfluencies. 
On the other hand, speech signals carry rich information beyond words via variations in timing, intonation, and loudness, i.e. in \emph{prosody}. Linguistic studies have shown that prosodic cues align with constituent structure \cite{Grosjean79}, signal disfluencies by marking the interruption point \cite{Shriberg94}, and help listeners resolve syntactic ambiguities \cite{Price1991}. Empirical evidence, however, has been mixed regarding the utility of prosody for constituency parsing. Most gains have been observed when sentence boundaries are unknown \cite{Kahn2012}, or with annotated prosodic labels \cite{Dreyer2007,Hale2006}. Most related to our current work, Tran et al.\ \cite{Tran2018} recently showed the benefit of using prosody in parsing within a sequence-to-sequence framework, proposing a convolutional neural network (CNN) as a mechanism to combine discrete word-level features with frame-level acoustic-prosodic features.

In this study, we extend the work in \cite{Tran2018} and \cite{Kitaev2018} to explore the utility of recent neural advances on spontaneous speech data, and compare the utility of prosody in read vs.\ spontaneous speech. Specifically, the goal of the current study is to answer the following questions: 
\begin{enumerate}[topsep=1pt,itemsep=0ex,partopsep=0ex,parsep=0.2ex]
\item Do contextualized word representations learned for written text also benefit spontaneous speech parsers? 
\item Does prosody improve further on top of the rich text information in neural parsers for spontaneous speech? 
\item How is the use of prosody affected by mismatch between read and spontaneous speech styles? 
\end{enumerate}


\begin{table*}[th]
\caption{Summary of datasets used}
\label{tab:data}
\centering
\begin{tabular}{lllll}
\toprule
Data & Style & Available material & Used for & \# sents \\ 
\midrule 
WSJ & news text & (gold) parses & train, dev (\textsection\ref{ssec:written-vs-speech}) & 40k \\ 
SWBD & conversational speech & audio, (gold) parses & train, dev, test (\textsection\ref{ssec:written-vs-speech}, \textsection\ref{ssec:prosody-vs-text}, \textsection\ref{ssec:train-test-mismatch}) & 96k \\ 
CSR & read news text & audio, (silver) parses & train (fine-tune), dev (\textsection\ref{ssec:prosody-vs-text}, \textsection\ref{ssec:train-test-mismatch}) & 8k \\ 
GT-N & read news/article text & audio, (gold) parses & test (\textsection\ref{ssec:train-test-mismatch}) & 6k (3k unique) \\ 
GT-SW & read version of SWBD & audio, (gold) parses & test/analysis (\textsection\ref{ssec:train-test-mismatch}) & 31 (13 unique) \\ 
\bottomrule
\end{tabular}
\end{table*}

%% file: models.tex
\section{Models}
\label{sec:models}
\begin{figure}[ht]
\centering
\includegraphics[scale=0.18]{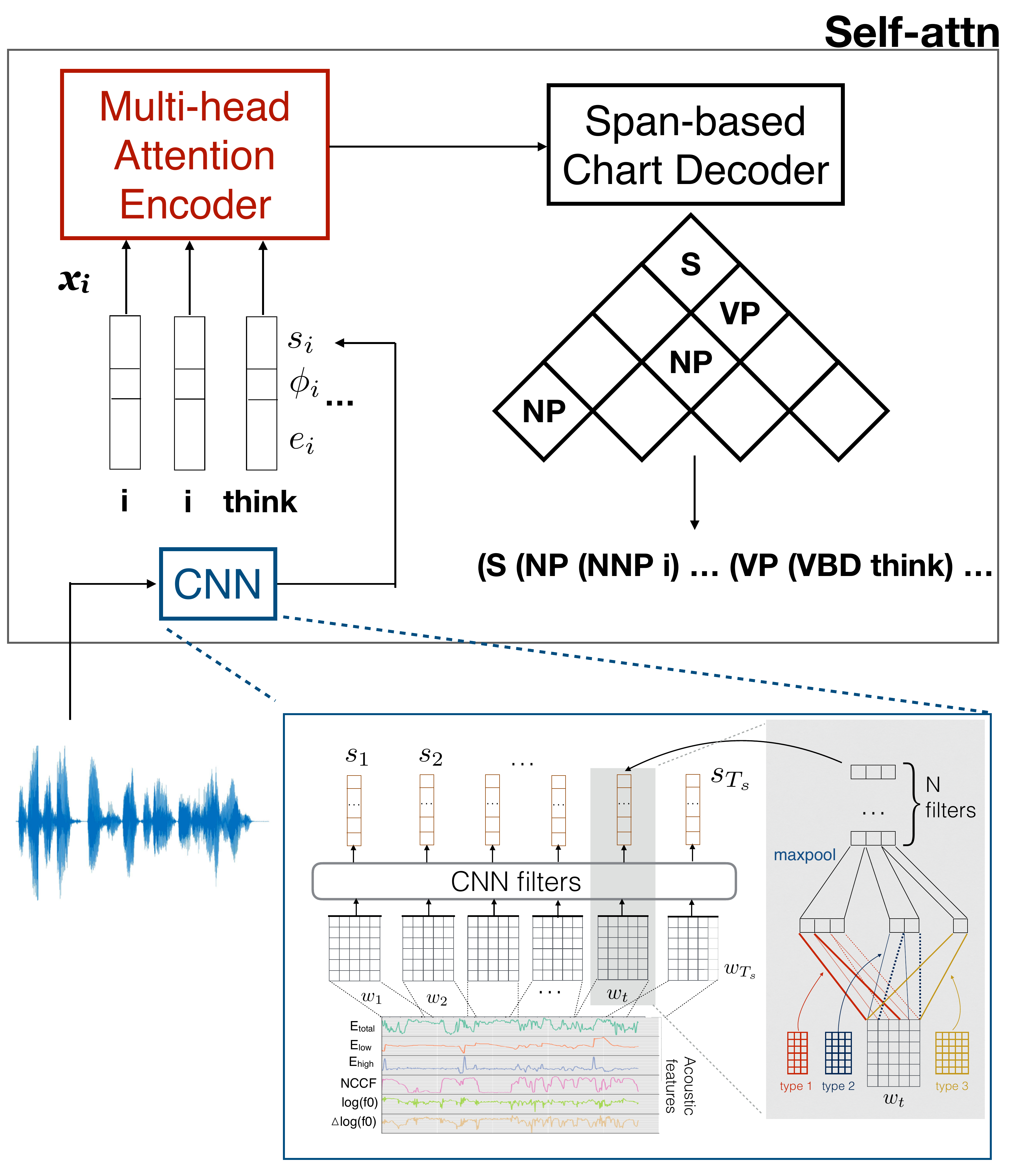}
\caption{Parser model overview, including: a CNN module for extracting prosodic features, a transformer encoder, and the chart decoder parser. Word-level input features include: word embeddings, pause/duration features, and CNN-based features.}
\label{fig:model-self-attn}
\end{figure}

We extend the self-attentive parser in \cite{Kitaev2018} to incorporate acoustic-prosodic information, which is learned through a convolutional neural network as in \cite{Tran2018}. Figure \ref{fig:model-self-attn} gives an overview of the model components. 

\subsection{Input representation}
\label{subsec:cnn}
Our parser model accepts a sequence of $T$ word-level features as inputs: $x_1, \cdots, x_{T}$, where $x_i = [e_i ; \phi_i ; s_i]$ is composed of word embeddings $e_i$, pause- and duration-based features $\phi_i$, and learned energy/pitch (E/f0) features $s_i$. The word embeddings $e_i$ can be (a) learned jointly with the parsing task, or (b) pretrained from external data and used as features, or (c) pretrained from external data and further fine-tuned jointly with the parsing task. In parsers that only use written text features (i.e. without acoustic-prosodic information), $x_i = e_i$. 

For acoustic-prosodic information, we follow the feature extraction process in \cite{Tran2018}. Briefly, $\phi_i$ contains word duration and pause features; $s_i$ is learned from energy (E) and pitch (f0) features via the CNN module. E/f0 frames are processed at the word level, where each sequence of frames corresponding to a time-aligned word (and potentially its surrounding context) is convolved with ($N$) filters of varying ($m$) sizes, resulting in $m\text{*}N$-dimensional speech features $s_i$. 

\subsection{Self-attentive parser}
\label{subsec:self-ann}
The self-attentive parser ({\bf Self-attn}) is composed of a multihead self-attention encoder, which follows the architecture of \cite{Vaswani2017}, and a span-based chart decoder, following the decoder from \cite{Stern2017,Gaddy2018}. Self-attn learns to predict a set of best-scoring labeled spans $(a, b, l)$, where $a, b \in [0, T]$ are position indices, and $l\in V_p$ is a label in the constituent label vocabulary $V_p$.

For each word $i$ in a sentence, the self-attentive encoder maps input $x_i$ to a query vector $q_i$, a key vector $k_i$, and a value vector $v_i$, which are used to compute the labeled span scores $s(a,b,l)$. 
Additionally, to capture the timing information without recurrent connections,  Self-attn uses positional embeddings $p_i$. These different types of inputs can be combined via simple addition or explicit factorization as detailed in \cite{Kitaev2018}. In our case, we extend the lexical-positional factorization in \cite{Kitaev2018} to lexical-positional-prosodic factorization. In particular, we learn separate key, query, and value mappings for each component of the input: $e_i, p_i$, and $[\phi_i, s_i]$.
Our implementation\footnote{\url{github.com/trangham283/prosody_nlp/tree/master/code/self_attentive_parser}} is based on the implementation in \cite{Kitaev2018} and includes the CNN module from \cite{Tran2018}. 

%% file: data-metrics.tex
\section{Data and Evaluation}
\label{sec:data-metrics}
\subsection{Datasets}
Table \ref{tab:data} summarizes the different datasets we used: some sets have both audio and parse trees available, while others have only either audio or parse trees.

We use two primary corpora for training and development: the Wall Street Journal ({\bf WSJ}) corpus of treebanked news articles  \cite{Marcus1999} and the Switchboard ({\bf SWBD}) corpus of telephone speech conversations \cite{Godfrey1993,Marcus1999}, which are the two standard corpora for constituency parsing studies on written text and conversational speech, respectively. SWBD includes audio files with time-aligned transcripts.

In order to train a parser with acoustic-prosodic features matched to the read speech style,
we use the common read subset of the CSR-I corpus  ({\bf CSR}) \cite{Garofolo1993}, which includes read Wall Street Journal sentences (but does not overlap with WSJ sentences). 
CSR is used to fine-tune a pre-trained SWBD parser (instead of training from scratch), since the corpus is much smaller than SWBD.
The Penn Phonetics Lab Forced Aligner (P2FA) \cite{p2fa} was used to get time alignments. 
Since the CSR sentences are not covered in the WSJ set, we used a pretrained state-of-the-art parser for written text \cite{Kitaev2018} to obtain silver trees. To verify the quality of the automatically parsed trees, we recruited two linguists to hand-correct a random subset of 100 trees. The annotator agreement is high: the F1 score between annotators' trees is 97.2\%. Among the 100 trees, both annotators confirmed that the parser got the perfect tree in 72 sentences, and the rest have minor errors.


To assess parser performance, we used the WSJ test set for written text, the SWBD test set for conversational speech, and two subsets of the GlobalTIMIT (GT) dataset \cite{GlobalTIMIT}.
{\bf GT-N} contains 3207 news sentences read by 50 speakers, some were read by multiple speakers, totaling 6k read sentences; {\bf GT-SW} contains the read version of 13 Switchboard sentences, read by 29 speakers, totaling 31 read sentences.\footnote{The number of read conversational sentences is limited, because we chose to use a standard corpus.} 
These sentences were selected from the Penn Treebank-3 corpus \cite{Marcus1999}, so they have gold-standard parse trees.

\subsection{Evaluation}
All models are evaluated using EVALB,\footnote{\url{nlp.cs.nyu.edu/evalb}} i.e.\ we report standard Parseval F1 scores. 
Because random seeds can lead to different results 
\cite{NilsGurevych2017}, we train and tune each model configuration initialized with 5 random seeds, and report the median prediction as our final result. Statistical significance was assessed using the paired bootstrap test as described in \cite{BBK2012:Significance}.

%% file: experiments.tex
\section{Experiments}
\label{sec:experiments}

\subsection{Do contextualized word representations learned for written text also benefit spontaneous speech parsers?}
\label{ssec:written-vs-speech}
To assess the impact of different types of text representations in parsing speech transcripts, we train and evaluate our parser on SWBD data, comparing several methods of using/learning word embeddings $e_i$. These embeddings can be learned jointly with the parsing task, or extracted from pretrained models and then used as features. For pretrained embeddings, we consider the following representations: GloVe \cite{Pennington2014glove} embeddings are learned from co-occurrence statistics and have little context information. The standard version (GloVe-Gigaword) was pretrained on a large corpus of 6B tokens (Wikipedia \& Gigaword 5). We additionally trained GloVe embeddings on a dataset with style more similar to spontaneous speech, the Fisher corpus \cite{fisher} and consider the effect of these embeddings on parsing (GloVe-Fisher). Contextualized embeddings such as ELMo \cite{Peters2018} and BERT \cite{Devlin2018} are recent neural models that have been pretrained on a large amount of written text data, capturing larger context information with language modeling auxiliary tasks via biLSTM (ELMo) or transformer networks (BERT). Both ELMo and BERT have been reported to benefit a variety of NLP tasks. 

Table \ref{tab:embedding-compare} presents results comparing embedding types on parsing SWBD test data. Using pretrained embeddings outperforms embeddings learned jointly with parsing, even though most pretrained models are on written text. 
Further, there is negligible difference between GloVe-Gigaword and the better matched GloVe-Fisher. This suggests that text features pretrained on large written text data do benefit parsing on speech transcripts.
Both contextualized models outperform GloVe models by a large margin (p-val $<$ 0.01), with BERT showing the best F1 scores, outperforming ELMo with statistical significance (p-val $<$ 0.01). This is consistent with results in other NLP tasks, confirming that contextualized embeddings are a powerful tool in a range of applications. All embeddings here are used as features, without further fine-tuning the embedding weights. We also ran several experiments where the embedding weights were jointly trained, but the results were worse, probably due to the large number of weights and the limited amount of speech transcripts.

\begin{figure*}[hbt]
  \centering
  \includegraphics[width=0.95\linewidth,trim={0 15.5cm 7cm 1.5cm},clip]{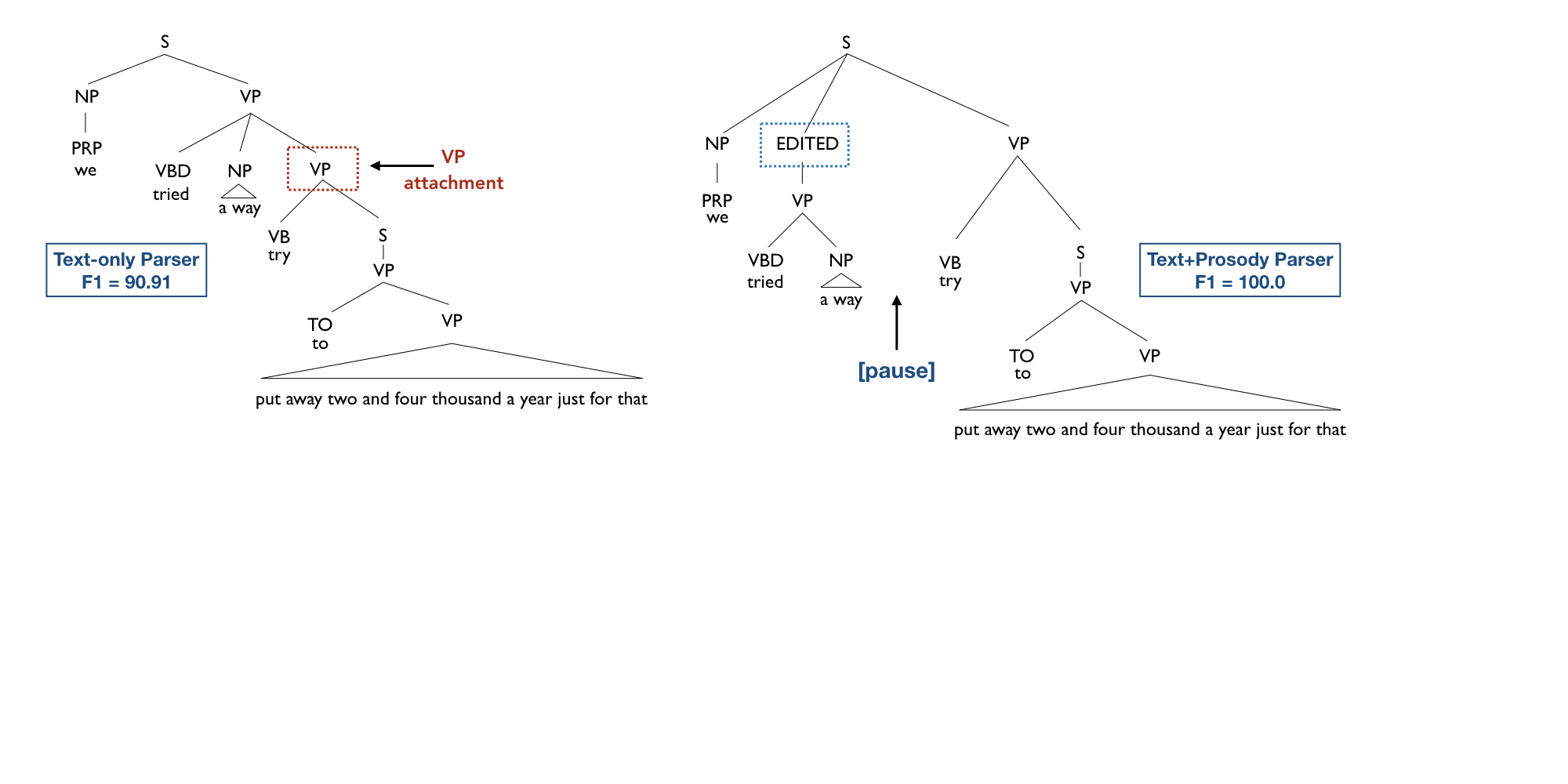}
  \caption{Predicted tree by a parser using only text (left) made a VP attachment error that the parser with prosody (right) avoided.}
  \label{fig:example}
\end{figure*}

\begin{table}[ht]
\centering
\caption{Parsing results on the SWBD dev set, using only text information, comparing different types of embeddings; all parsers were trained on the SWBD train set. Differences between BERT vs.\ ELMo, and those between BERT/ELMo vs.\ others are statistically significant with p-val $< 0.01$.}
\label{tab:embedding-compare}
\begin{tabular}{ll}
\toprule
Embedding & F1 \\ \midrule 
Learned & 90.98 \\ 
GloVe - Fisher & 91.04 \\ 
GloVe - Gigaword \cite{Pennington2014glove}& 91.17 \\ 
ELMo \cite{Peters2018} & 92.69 \\ 
BERT \cite{Devlin2018} & 93.24 \\ 
\bottomrule
\end{tabular}
\end{table}

Similar to comparing different types of embeddings, we also assess the effect of using different datasets on parsing speech transcripts. Table \ref{tab:treebank-compare} presents these results. Unsurprisingly, simply training on written text data performs poorly on speech transcripts. Training on additional text-only data (SWBD+WSJ) provides marginal improvement in parsing conversational speech, suggesting that substantial benefit can be obtained with pretrained embeddings, but the dataset for the main task still requires a style match. 

\begin{table}[ht]
\centering
\caption{Parsing result on the SWBD dev set, using only text information, comparing different types of training data. The differences between SWBD and SWBD+WSJ are not significant.}
\label{tab:treebank-compare}
\begin{tabular}{lll}
\toprule
Trained on & ELMo & BERT \\ \midrule
WSJ & 76.00 & 77.45 \\ 
SWBD & 92.69 & 93.24 \\ 
SWBD + WSJ & 92.70 & 93.40 \\ 
\bottomrule
\end{tabular}
\end{table}

\subsection{Does prosody improve further on top of the rich text information in neural parsers for spontaneous speech?}
\label{ssec:prosody-vs-text}

For this question, we only consider the two best-performing models on text-only data: Self-attn with ELMo vs.\ BERT. Table \ref{tab:text-vs-pros} presents the results on SWBD test set, separating results by fluent vs.\ disfluent (sentences with EDITED and/or INTJ nodes) subsets of sentences. 

\begin{table}[htb]
\caption{Parsing results on the SWBD test set 
(3823 disfluent + 2078 fluent sentences): using only text information vs.\ adding acoustic-prosodic features. Comparing text+prosody and text models, statistical significance is denoted as: (*) p-val $<$ 0.02; (\textdagger) p-val $<$ 0.05. 
}\label{tab:text-vs-pros}
\centering
\begin{tabular}{lllll}
\toprule
Model & Embedding & all & disfluent & fluent  \\ \midrule
\multirow{2}{*}{text} & ELMo & 92.47 & 91.48 & 94.64 \\
& BERT & 92.86 & 91.92 & 94.89  \\ \midrule
\multirow{2}{*}{+prosody} & ELMo & 92.68* & 91.68* & 94.86\textdagger  \\
& BERT & 93.04* & 92.06 & 95.16* \\ 
\bottomrule
\end{tabular}
\end{table}

Comparing text-only and text+prosody models, prosody helps in both ELMo and BERT. ELMo results are consistent with previous work \cite{Tran2018}: most gains seem to be from disfluent sentences. For BERT, the gains are statistically significant in fluent sentences, but not in disfluent ones. 
Comparing BERT and ELMo models, BERT-text improves over ELMo-text with p-val $<$ 0.05 in disfluent sentences and overall, but not in fluent sentences. This is likely why BERT-prosody does not improve over BERT-text with statistical significance in disfluent sentences, since BERT-text itself is already good. BERT-prosody improves over ELMo-prosody in all cases with p-val $<$ 0.05.

\begin{table}[ht]
\centering
\caption{Test set F1 scores for different sentence lengths. Prosody shows the most benefit in long sentences.}
\label{tab:sent-len}
\begin{tabular}{llcccc}
\toprule
 & & & \multicolumn{3}{c}{Sentence lengths (\# sents)} \\ \cmidrule{4-6}
Embedding & Model & & \begin{tabular}[c]{@{}c@{}}{[}0, 5{]}\\ (2885)\end{tabular} & \begin{tabular}[c]{@{}c@{}}{[}6, 10{]}\\ (1339)\end{tabular} & \begin{tabular}[c]{@{}c@{}}{[}11, -{]}\\ (1677)\end{tabular} \\ \cmidrule{1-2}\cmidrule{4-6}
\multirow{2}{*}{ELMo} & text && 96.64 & 96.33 & 90.53 \\ 
 & +prosody && 96.65 & 96.43 & 90.81 \\ \cmidrule{1-2}\cmidrule{4-6}
\multirow{2}{*}{BERT} & text && 96.51 & 96.53 & 91.07 \\ 
 & +prosody && 96.63 & 96.67 & 91.30 \\ 
 \bottomrule
\end{tabular}
\end{table}

Table \ref{tab:sent-len} shows the parse scores for subsets of sentences grouped by length. For both ELMo and BERT, prosody benefits parsing more for longer sentences than short ones. 

\begin{table}[ht]
\centering
\caption{Percentage of error reduction counts from text to text+prosody models (first 2 columns) and from ELMo to BERT models (last 2 columns).}
\label{tab:error-type-reduction}
\begin{tabular}{lccccc}
\toprule
\multirow{2}{*}{Error Type} & \multicolumn{2}{c}{$\Delta$(+pros, text)} && \multicolumn{2}{c}{$\Delta$(BERT, ELMo)} \\ 
\cmidrule(l){2-3}\cmidrule(r){5-6}
 & ELMo & BERT && text & +pros \\ \midrule
Co-ordination & -1.0 & -5.1 && 18.2 & 14.9 \\ 
PP Attach. & 1.2 & 1.0 && 1.2 & 1.0 \\ 
NP Attach. & -7.5 & 0.0 && 6.0 & 12.5 \\ 
VP Attach. & 19.2 & 19.6 && -7.7 & -7.1 \\ 
Clause Attach. & 8.3 & -8.1 && 11.4 & -4.4 \\ 
Mod. Attach. & 7.9 & -1.4 && 11.8 & 3.0 \\ 
NP Internal & 2.7 & 7.0 && 6.5 & 10.6 \\
1-Word Phrase & 5.2 & 2.3 && -3.5 & -6.6 \\ 
Different Label & 1.0 & 7.3 && -2.4 & 4.1 \\ 
\bottomrule
\end{tabular}
\end{table}

Similar to \cite{Tran2018}, we also analyze parse error types each parser makes or improves on. We use the Berkeley Parse Analyzer \cite{Kummerfeld2012} to categorize the common error types in constituency parsing. Table \ref{tab:error-type-reduction} shows the relative error reduction when using prosody vs.\ using only text, and similarly when using BERT vs.\ ELMo. For both ELMo and BERT, VP attachment errors are most reduced when using prosody. Figure \ref{fig:example} shows an example sentence where prosodic features help avoid the attachment error made by the parser using only text features. 
Cases where prosody seems to hurt BERT (Coordination, Clause Attachment, and possibly Modifier Attachment) are contexts where the text-only BERT and ELMo models have the greatest difference. For the main case where prosody hurts ELMo (NP Attachment), there is no benefit to BERT. These may simply be contexts where there is little need for prosody given well-trained text models.  


\subsection{How is the use of prosody affected by mismatch between read and spontaneous speech styles?} 
\label{ssec:train-test-mismatch}

For this experiment, we only consider the models with BERT. 
Table~\ref{tab:mismatch-train-test} presents parsing results in mismatched tuning-testing conditions. 
In all settings, training on conversational speech degrades results on read speech minimally, but training on read speech degrades results on conversational speech significantly. Further, prosody consistently helps when the parser is trained on conversational speech, both when testing with matched and mismatched styles. This suggests that conversational speech data is more useful for general purpose parser training, likely because of the diversity in prosodic characteristics available in spontaneous speech. 

When testing on conversational speech (SWBD column), the biggest effect of mismatch is associated with the word sequence; the degradation from prosody mismatch seems to have a smaller but still significant impact (p-val $<0.05$). 
However, when testing on read news (GT-N column), the BERT model with prosody tuned on read speech sees a performance gain (p-val $<0.01$). 
These results are consistent with the hypothesis that use of prosody differs in read vs.\ conversational speech, i.e. the style mismatch is both in terms of words and acoustic cues.

\begin{table}[ht]
\centering
\caption{Results for mismatched tuning-testing conditions: conversational (C) vs.\ read (R) vs.\ read conversational transcripts (RC). Comparing the improvement of text+prosody over text models, statistical significance is denoted as: (*) p-val $<$ 0.02.}
\label{tab:mismatch-train-test}
\begin{tabular}{llllc}\toprule
 & & \multicolumn{3}{c}{Test data} \\ \cmidrule{3-5}
\begin{tabular}[c]{@{}l@{}}Tuning\\ data\end{tabular} & \multirow{2}{*}{Model} & \begin{tabular}[c]{@{}c@{}}SWBD \\ (C)\end{tabular} & \begin{tabular}[c]{@{}c@{}}GT-N \\ (R)\end{tabular} & \begin{tabular}[c]{@{}c@{}}GT-SW\\ (RC)\end{tabular} \\ \midrule
SWBD (C) & text & 92.86 & 92.35 & 97.96 \\
CSR (R) & text & 80.60 & 93.92 & 91.38 \\
SWBD (C) & +prosody & 93.04* & 92.58* & 97.96 \\
CSR (R) & +prosody & 80.36 & 94.17* & 90.29 \\
\bottomrule
\end{tabular}
\end{table}

To further explore this question, we ran experiments on the GT-SW sentences. The results in Table~\ref{tab:mismatch-train-test} (GT-SW column) are anecdotal but consistent with the other results. On these sentences, with text-only models, further tuning on read style data degrades performance significantly. For the parsers using prosody, the version trained on spontaneous speech seems to be able to handle the read version of Switchboard sentences, but the one fine-tuned on read text further degrades.
It may be that the prosody associated with reading conversation transcripts is not like that associated with reading more formal written text.


%% file: conclusion.tex
\section{Conclusion}
\label{sec:conclusion}
We show that neural architectures, in particular contextualized embeddings pretrained on large written text (ELMo, BERT), improve constituency parsing on conversational speech transcripts. The use of prosody results in further improvements overall, especially in longer sentences and in reducing attachment errors. Assessing the utility of prosody in different speaking styles, we found that parsers trained with spontaneous prosody are consistently useful, improving over their text-only counterparts when testing on both conversational and read (mismatched) speech. Fine-tuning such parsers on read speech improves results when testing on the same read style, but degrades significantly on spontaneous speech. This suggests that conversational speech data is more useful for general parser training.

%% file: thanks.tex
\section{Acknowledgements}
This work was funded in part by the US National Science Foundation (NSF), grant IIS-1617176. 
Any opinions, conclusions or recommendations expressed in this material are those of the authors and do not necessarily reflect the views of the NSF.

%% file: main_parsing_speech.bbl
\begin{thebibliography}{10}
\providecommand{\url}[1]{#1}
\csname url@samestyle\endcsname
\providecommand{\newblock}{\relax}
\providecommand{\bibinfo}[2]{#2}
\providecommand{\BIBentrySTDinterwordspacing}{\spaceskip=0pt\relax}
\providecommand{\BIBentryALTinterwordstretchfactor}{4}
\providecommand{\BIBentryALTinterwordspacing}{\spaceskip=\fontdimen2\font plus
\BIBentryALTinterwordstretchfactor\fontdimen3\font minus
  \fontdimen4\font\relax}
\providecommand{\BIBforeignlanguage}[2]{{%
\expandafter\ifx\csname l@#1\endcsname\relax
\typeout{** WARNING: IEEEtran.bst: No hyphenation pattern has been}%
\typeout{** loaded for the language `#1'. Using the pattern for}%
\typeout{** the default language instead.}%
\else
\language=\csname l@#1\endcsname
\fi
#2}}
\providecommand{\BIBdecl}{\relax}
\BIBdecl

\bibitem{Marcus1999}
M.~P. Marcus, B.~Santorini, M.~A. Marcinkiewicz, and A.~Taylor,
  \emph{Treebank-3 LDC99T42}, Linguistic Data Consortium, 1999.

\bibitem{Vinyals2015}
O.~Vinyals, L.~Kaiser, T.~Koo, S.~Petrov, I.~Sutskever, and G.~Hinton,
  ``Grammar as a {F}oreign {L}anguage,'' in \emph{Proc. NIPS}, 2015.

\bibitem{Stern2017}
M.~Stern, J.~Andreas, and D.~Klein, ``A minimal span-based neural constituency
  parser,'' in \emph{Proc. ACL}, 2017, pp. 818--827.

\bibitem{Shen2018}
Y.~Shen, Z.~Lin, A.~P. Jacob, A.~Sordoni, A.~Courville, and Y.~Bengio,
  ``Straight to the tree: Constituency parsing with neural syntactic
  distance,'' in \emph{Proc. ACL}, 2018, pp. 1171--1180.

\bibitem{Vilares2018}
C.~G{\'o}mez-Rodr{\'i}guez and D.~Vilares, ``Constituent parsing as sequence
  labeling,'' in \emph{Proc. EMNLP}, 2018, pp. 1314--1324.

\bibitem{Kitaev2018}
N.~Kitaev and D.~Klein, ``Constituency parsing with a self-attentive encoder,''
  in \emph{Proc. ACL}, 2018, pp. 2676--2686.

\bibitem{Vaswani2017}
A.~Vaswani, N.~Shazeer, N.~Parmar, J.~Uszkoreit, L.~Jones, A.~N. Gomez,
  L.~Kaiser, and I.~Polosukhin, ``Attention is all you need,'' in
  \emph{Advances in Neural Information Processing Systems 30}, I.~Guyon, U.~V.
  Luxburg, S.~Bengio, H.~Wallach, R.~Fergus, S.~Vishwanathan, and R.~Garnett,
  Eds., 2017, pp. 5998--6008.

\bibitem{Peters2018}
M.~Peters, M.~Neumann, M.~Iyyer, M.~Gardner, C.~Clark, K.~Lee, and
  L.~Zettlemoyer, ``Deep contextualized word representations,'' in \emph{Proc.
  NAACL}, 2018, pp. 2227--2237.

\bibitem{Devlin2018}
J.~Devlin, M.-W. Chang, K.~Lee, and K.~Toutanova, ``{BERT}: Pre-training of
  deep bidirectional transformers for language understanding,'' in \emph{Proc.
  NAACL}, 2019, pp. 4171--4186.

\bibitem{Grosjean79}
F.~Grosjean, L.~Grosjean, and H.~Lane, ``The patterns of silence: {P}erformance
  structures in sentence production,'' \emph{Cognitive Psychology}, 1979.

\bibitem{Shriberg94}
E.~Shriberg, ``Preliminaries to a theory of speech disfluencies,'' Ph.D.
  dissertation, Department of Psychology, University of California, Berkeley,
  CA, 1994.

\bibitem{Price1991}
P.~Price, M.~Ostendorf, S.~Shattuck-Hufnagel, and C.~Fong, ``The {U}se of
  {P}rosody in {S}yntactic {D}isambiguation,'' in \emph{Proc.~Workshop on
  Speech and Natural Language}, 1991.

\bibitem{Kahn2012}
J.~G. Kahn and M.~Ostendorf, ``Joint reranking of parsing and word recognition
  with automatic segmentation,'' \emph{Computer Speech \& Language}, 2012.

\bibitem{Dreyer2007}
M.~Dreyer and I.~Shafran, ``Exploiting prosody for {PCFG}s with latent
  annotations,'' in \emph{Proc. Interspeech}, 2007.

\bibitem{Hale2006}
J.~Hale, I.~Shafran, L.~Yung, B.~Dorr, M.~Harper, A.~Krasnyanskaya, M.~Lease,
  Y.~Liu, B.~Roark, M.~Snover, and R.~Stewart, ``{PCFG}s with {S}yntactic and
  {P}rosodic {I}ndicators of {S}peech {R}epairs,'' in \emph{Proc. COLING-ACL},
  2006.

\bibitem{Tran2018}
T.~Tran, S.~Toshniwal, M.~Bansal, K.~Gimpel, K.~Livescu, and M.~Ostendorf,
  ``Parsing speech: a neural approach to integrating lexical and
  acoustic-prosodic information,'' in \emph{Proc. NAACL}, 2018, pp. 69--81.

\bibitem{Gaddy2018}
D.~Gaddy, M.~Stern, and D.~Klein, ``{What's Going On in Neural Constituency
  Parsers? An Analysis},'' in \emph{Proc. NAACL}, 2018, pp. 999--1010.

\bibitem{Godfrey1993}
J.~J. Godfrey and E.~Holliman, \emph{{Switchboard-1 {R}elease 2 LDC97S62}},
  Linguistic Data Consortium, 1993.

\bibitem{Garofolo1993}
J.~S. Garofolo, D.~Graff, D.~Paul, and D.~Pallett, \emph{{CSR-I (WSJ0) Complete
  LDC93S6A}}, Linguistic Data Consortium, 1993.

\bibitem{p2fa}
J.~Yuan and M.~Liberman, ``Speaker identification on the {SCOTUS} corpus,'' in
  \emph{Proc. Acoustics}, 2008.

\bibitem{GlobalTIMIT}
N.~Chanchaochai, C.~Cieri, J.~Debrah, H.~Ding, Y.~Jiang, S.~Liao, M.~Liberman,
  J.~Wright, J.~Yuan, J.~Zhan, and Y.~Zhan, ``Globaltimit: Acoustic-phonetic
  datasets for the world's languages,'' in \emph{Proc.\ Interspeech}, 2018, pp.
  192--196.

\bibitem{NilsGurevych2017}
N.~Reimers and I.~Gurevych, ``Reporting score distributions makes a difference:
  Performance study of lstm-networks for sequence tagging,'' in \emph{Proc.
  EMNLP}, 2017, pp. 338--348.

\bibitem{BBK2012:Significance}
T.~Berg-Kirkpatrick, D.~Burkett, and D.~Klein, ``{An Empirical Investigation of
  Statistical Significance in NLP},'' in \emph{Proc. EMNLP}, 2012.

\bibitem{Pennington2014glove}
J.~Pennington, R.~Socher, and C.~D. Manning, ``Glove: Global vectors for word
  representation,'' in \emph{Empirical Methods in Natural Language Processing
  (EMNLP)}, 2014, pp. 1532--1543.

\bibitem{fisher}
C.~Cieri, D.~Graff, O.~Kimball, D.~Miller, and K.~Walker, ``Fisher english
  training speech part 1 transcripts, {LDC}2004{T}19,'' Web Download, 2004.

\bibitem{Kummerfeld2012}
J.~K. Kummerfeld, D.~Hall, J.~R. Curran, and D.~Klein, ``Parser {S}howdown at
  the {W}all {S}treet {C}orral: {A}n {E}mpirical {I}nvestigation of {E}rror
  {T}ypes in {P}arser {O}utput,'' in \emph{Proc. EMNLP}, 2012.

\end{thebibliography}
